\documentclass[letterpaper]{article} 
\usepackage{aaai2026}  
\usepackage{times}  
\usepackage{helvet}  
\usepackage{courier}  
\usepackage[hyphens]{url}  
\usepackage{graphicx} 
\usepackage{natbib}  
\usepackage{caption} 
\usepackage{algorithm}
\usepackage{algorithmic}
	
\usepackage{amssymb}
\usepackage{multirow}
\usepackage{xspace}
\usepackage{url}
\usepackage{bm}
\usepackage{pifont}
\usepackage{makecell}	
\usepackage{booktabs}

\usepackage{newfloat}
\usepackage{listings}
\DeclareCaptionStyle{ruled}{labelfont=normalfont,labelsep=colon,strut=off} 
\floatstyle{ruled}
\newfloat{listing}{tb}{lst}{}
\floatname{listing}{Listing}
\title{CoopDiff: Anticipating 3D Human-object Interactions via Contact-consistent Decoupled Diffusion}
\author {
    Xiaotong Lin\textsuperscript{\rm 1},
    Tianming Liang\textsuperscript{\rm 1},
    Jian-Fang Hu\textsuperscript{\rm 1}*,
    Kun-Yu Lin\textsuperscript{\rm 2},
    Yulei Kang\textsuperscript{\rm 1},
    Chunwei Tian\textsuperscript{\rm 3},
    Jianhuang Lai\textsuperscript{\rm 1},
    Wei-Shi Zheng\textsuperscript{\rm 1},
}
\affiliations {
    \textsuperscript{\rm 1} Sun Yat-sen University,
    \textsuperscript{\rm 2} The University of Hong Kong,
    \textsuperscript{\rm 3} Harbin Institute of Technology\\
    linxt29@mail2.sysu.edu.cn, hujf5@mail.sysu.edu.cn
}

\usepackage{bibentry}

\begin{document}

\maketitle

\begin{abstract}
3D human-object interaction (HOI) anticipation aims to predict the future motion of humans and their manipulated objects, conditioned on the historical context. Generally, the articulated humans and rigid objects exhibit different motion patterns, due to their distinct intrinsic physical properties. However, this distinction is ignored by most of the existing works, which intend to capture the dynamics of both humans and objects within a single prediction model.
In this work, we propose a novel contact-consistent decoupled diffusion framework \textbf{CoopDiff}, which employs two distinct branches to decouple human and object motion modeling, with the human-object contact points as shared anchors to bridge the motion generation across branches. The human dynamics branch is aimed to predict highly structured human motion, while the object dynamics branch focuses on the object motion with rigid translations and rotations. 
These two branches are bridged by a series of shared contact points with consistency constraint for coherent human-object motion prediction. To further enhance human-object consistency and prediction reliability, we propose a human-driven interaction module to guide object motion modeling. Extensive experiments on the BEHAVE and Human-object Interaction datasets demonstrate that our CoopDiff outperforms state-of-the-art methods.
\end{abstract}


\section{Introduction}
\label{sec:intro}

Given a historical sequence that captures the dynamics of humans manipulate or interact with objects, 3D human-object interaction (HOI) anticipation aims to precisely predict future motions for both human and object, and ensures their interactions coherent and physically plausible. Recently, this problem has received increasing interest due to its significant real-world application value in robotics \cite{gu2024interactive,lee2024commonsense}, animation \cite{song2023automatic}, augmented reality \cite{hu2021fixationnet,li2024favor}, and embodied AI \cite{padmakumar2022teach}.

Previous methods for this task \cite{ghosh2023imos,li2024task,taheri2022goal,wu2022saga} are primarily designed to predict the human dynamics rather than human-object dynamics, with an assumption that the objects remain static or relatively static to human hands. This largely oversimplifies the interactions between humans and objects. Recently, InterDiff \cite{xu2023interdiff} explores 3D HOI anticipation in a more reasonable setting, aiming to capture the dynamics of both humans and objects simultaneously within a \textit{single} prediction model. However, all these methods overlook the significant difference in dynamics between humans and objects, often leading to less precise and realistic results, e.g., interpenetration and object floating (please refer to Figure \ref{fig:BEHAVE_vis_ST} for the visualization results).

\begin{figure}[!t]
\centering
\vspace{-3mm}
\includegraphics[width=0.95\linewidth]{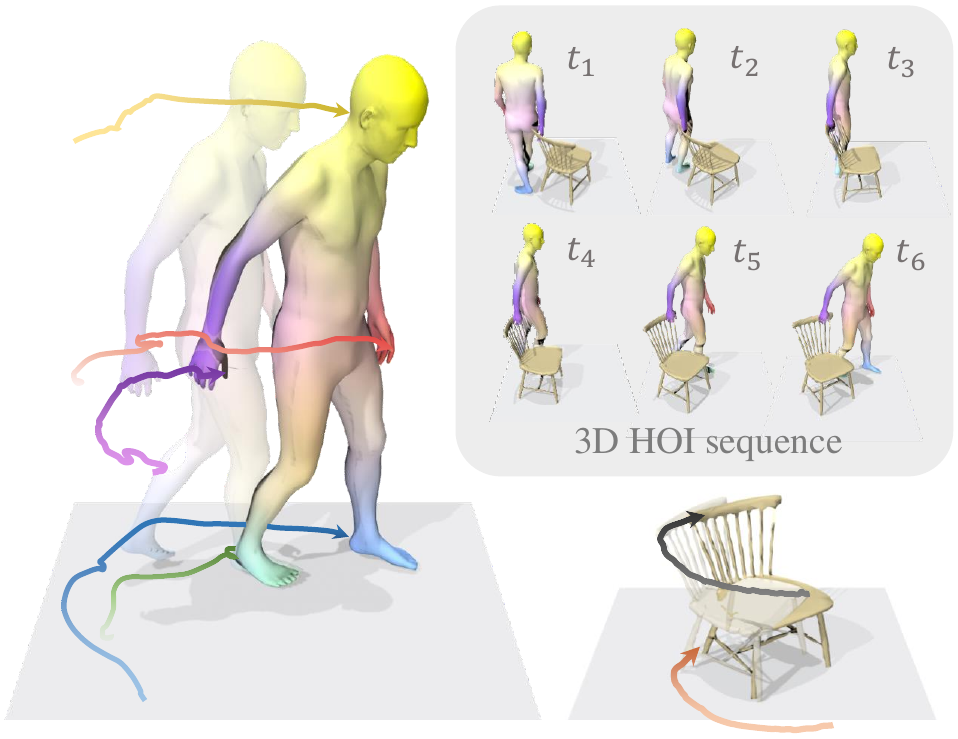}
\vspace{-2mm}
\caption{Visualization results illustrating the \textit{difference} between human and object dynamics in HOI. We visualize the motions of various human joints and object parts. Lines with varied colors indicate the trajectories of joints corresponding to human body and object. As shown, the articulated human body is highly structured with diverse movements across joints, whereas object motion typically involves rigid translations or rotations. }
\label{fig:intro}
\end{figure}

Taking the interaction ``A man pulls the chair around" (see Figure \ref{fig:intro}) as an example, we observe that humans
and objects demonstrate distinct dynamic patterns in terms of kinematic modes and physical properties. The articulated human body often exhibits diverse movement patterns across joints, whereas objects typically involve rigid translations or rotations when manipulated by humans through contact points. These motivate us to model the human and object dynamics from different perspectives.

\begin{figure*}[!ht]
\centering
\includegraphics[width=0.89\linewidth]{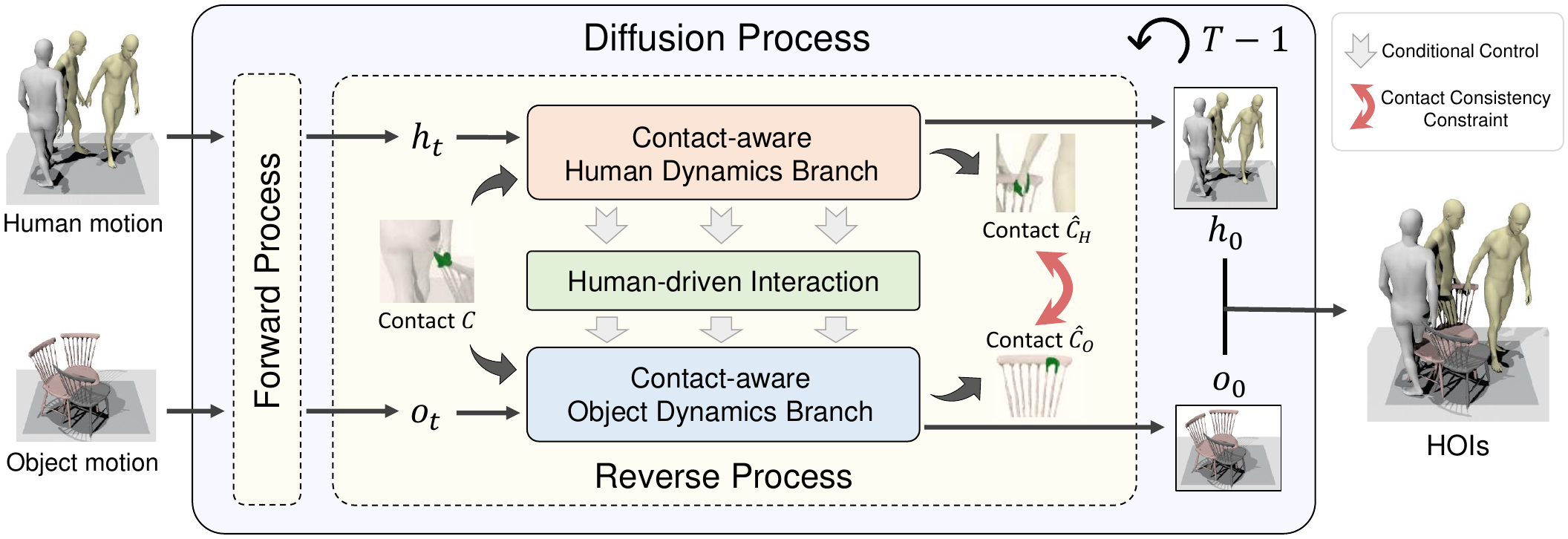}
\vspace{-1mm}
\caption{Illustration of CoopDiff framework. CoopDiff employs a dual-branch diffusion to separately model the distinct motion patterns of humans and objects. For both branches, we additionally integrate contact points as shared anchors and feed them into branches for additional contact prediction. These contact predictions across branches are aligned with a consistency constraint to ensure coherent human-object dynamics modeling. To further enhance coherence between human-object dynamics, we devise a human-driven interaction module that incorporates human dynamics as conditional control to object dynamics modeling.}
\label{fig:Architecture}
\end{figure*}

In this paper, we propose a novel \textbf{Co}ntact-c\textbf{o}nsistent decou\textbf{p}led \textbf{Diff}usion model (CoopDiff), which employs a dual-branch diffusion to separately model the distinct patterns of human and object motions while integrating contact points to ensure consistency between branches. Specifically, our proposed CoopDiff consists of a human dynamics branch to predict highly structured human motion, and an object dynamics branch to predict object dynamics with rigid translations and rotations. To facilitate coherent dynamics modeling across branches, we introduce contact points as shared anchors and feed them into each branch for additional contact prediction under different motion patterns. In this way, the contact points deftly bridge the two individual branches, providing a consistency constraint to ensure the alignment between human and object motions. 
To further improve the alignment and ensure reliable prediction, we develop a human-driven interaction module to guide object dynamics modeling with human dynamics, considering that human typically takes the dominant role as the force-exerting agent on objects. 
Combining these designs, our CoopDiff can effectively capture the distinct dynamics of humans and objects, generating more precise and realistic 3D HOI motions.

Extensive experiments on BEHAVE \cite{bhatnagar2022behave} and Human-object Interaction datasets \cite{wan2022learn} show that our framework, with decoupled human-object modeling and contact consistency, outperforms state-of-the-art methods by a large margin.
Our main contributions are summarized as follows:
\begin{itemize}
\item We present a novel contact-consistent decoupled framework for 3D HOI anticipation, which 
separately captures distinct motion patterns of humans and objects by a dual-branch diffusion, and ensures coherent dynamics modeling with contact consistency;

\item We devise a human-driven interaction module to enhance human-object interaction modeling, which empirically mitigates unrealistic interactions;

\item Extensive experiments on two benchmark datasets demonstrate that our framework consistently outperforms the state-of-the-art methods.
\end{itemize}





\section{Related Work}
\label{sec:related}

\noindent\textbf{3D Human-Object Interaction (HOI) Anticipation.}
Existing research in 3D human-object interaction (HOI) anticipation generally falls into two categories: skeleton-based and mesh-based approaches. While skeleton-based methods \cite{corona2020context,yan2024forecasting,hu2024hoimotion,razali2023action,wan2022learn} focus mainly on joint accuracy; mesh-based approaches \cite{xu2023interdiff,zhou2024gears,diller2024cg,zeng2025chainhoi,xu2025intermimic,xue2025guiding} additionally emphasize interaction realism, addressing challenges like object floating and mesh penetration, which makes them more applicable to real-world scenarios. 
Early mesh-based studies \cite{jiang2021hand,kwon2021h2o,hasson2019learning,zhang2021interacting} primarily investigate the interactions between hands and objects. Recently, the whole-body HOI anticipation \cite{ghosh2023imos,kulkarni2024nifty,taheri2022goal,wu2022saga,li2024task} has emerged and attracted increasing attention. Despite the progress, most works remain constrained to humans grasping small, static objects with relatively fixed shapes, overlooking highly complicated, dynamic interactions present in real-world scenarios. Recently, InterDiff \cite{xu2023interdiff} reformulates this task for dynamic and diverse objects in more realistic scenarios. It leverages a diffusion model to jointly forecast the dynamics of humans and objects, with a correction module to rectify implausible interactions. However, these methods neglect the distinct motion patterns of humans and objects, either by employing a single prediction model for both or concentrating mainly on human motion. In this work, we address this issue with a novel decoupled modeling framework, aiming to accurately capture the distinct dynamic patterns of humans and objects.


\vspace{0.5em}
\noindent\textbf{Decoupled Human-object Modeling.} 
To generate interactions between whole-body movements and dynamic objects, several attempts \cite{peng2025hoi,nam2024joint,cao2024avatargo} have been made to separately model the dynamics of humans and objects. For instance, in text-driven HOI synthesis, HOI-Diff \cite{peng2025hoi} employs two separate Transformers to model humans and objects, with a mutual cross-attention facilitating information exchange. Nevertheless, these methods have two main shortcomings: i) most approaches treat the dynamics of humans and objects equally, neglecting that humans typically dominate interactions as a manipulator in HOI; ii) They tend to consider contact points as a separate category for prediction, forecasting them separately and using them only for post-processing correction between humans and objects. This could result in suboptimal utilization of contact information. In contrast, we treat contacts as shared anchors to accommodate distinct motions of humans and objects, with a human-driven interaction module to enhance the interaction modeling.

\section{Method}
\label{sec:method}

\noindent\textbf{Overview.}
Based on the historical HOI sequence, CoopDiff aims to predict the future unobserved dynamics of both humans and the objects they manipulate. Figure \ref{fig:Architecture} illustrates the overall framework. Our CoopDiff employs a dual-branch diffusion to separately capture the distinct dynamics of humans and objects, with their shared contact points integrated into both branches as a bridge to facilitate coherent dynamics modeling. Specifically, the human branch focuses on capturing the highly structured human dynamics, by iterative denoising to recover clean motion and future contact conditioned on history (Sec. \ref{sec:branch_h}). While the object branch is designed to model rigid motion, through a similar denoising process to generate clean object motion along with contact (Sec. \ref{sec:branch_o}). We then align these contact predictions across branches with a consistency constraint, ensuring coherent human-object dynamics modeling (Sec. \ref{sec:CCC}). To further enhance dynamics coherence across branches, we devise a human-driven interaction module that incorporates knowledge gained in human dynamics learning into object dynamics modeling, treating humans as manipulators on objects to closely align with real-world scenarios (Sec. \ref{subsec:CSIM}). 




\begin{figure}[!t]
\centering
\vspace{-0.5mm}
\includegraphics[width=0.95\linewidth]{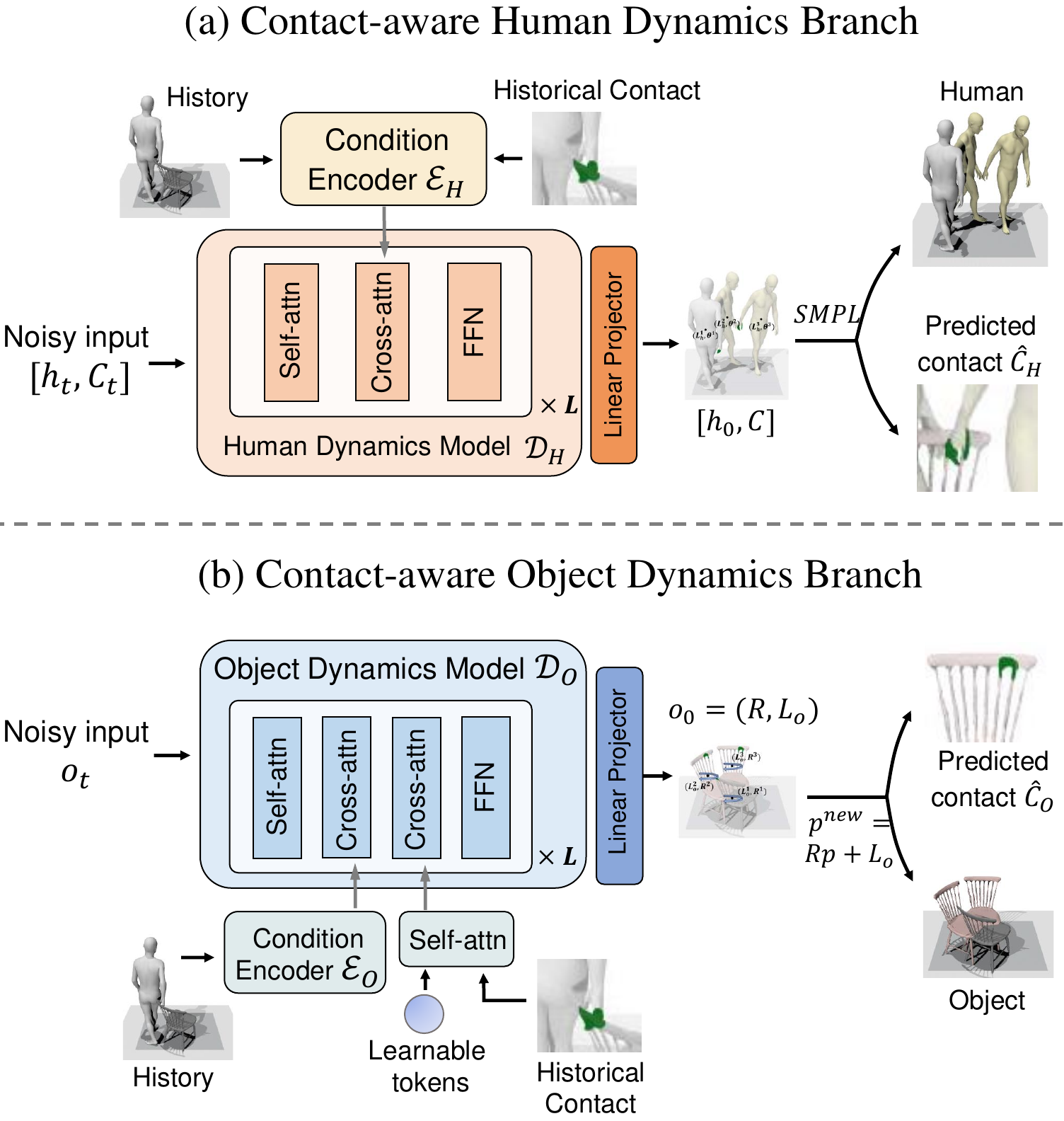}
\vspace{-2.2mm}
\caption{Contact-aware human and object dynamics branch.}
\label{fig:CCL}
\end{figure}

\subsection{Data Representation}
We denote a 3D HOI sequence with $T_p$ past frames and $T_f$ future frames as $\mathcal{S}^{1:T_p+T_f}$, where each frame is composed of human pose state and object pose state. The human pose state $\bm{h}$ is represented as the 3D positions $\bm{L}_h$ and 6D continuous rotations $\bm{\theta}$; then the human mesh can be reconstructed from these pose and body shape parameters using SMPL model \cite{loper2023smpl,romero2022embodied}. For object pose state $\bm{o}$, we represent it using 3D positions $\bm{L}_o$ and relative rotations $\bm{R}$. The 3D positions correspond to the object centroids, and the relative rotations are defined with respect to the given object point cloud $\bm{P}$, such that the vertices on the object surface are expressed as $\bm{V}=\bm{R} \cdot \bm{P} + \bm{L}_o$. Regarding the contact points, we define them as the shared points on both surfaces of the interacting human and object. Considering that the contact points could distributed densely within specific body regions (e.g., left hand), rather than retaining all of them inefficiently, we select to extract representative points from each body region for computational efficiency.  
Specifically, we first categorize contact points into $N$ groups based on their distance to $N$ human joints, as denoted by $\{\bm{g}_1, \bm{g}_2, \dots, \bm{g}_N\}$. In this way, each contact point is assigned to one of the $N$ body regions. A binary contact label $\mathcal{M} \in \mathcal{R}^N$ is defined to indicate whether the groups (body regions) contain any contact points, i.e., $\mathcal{M}_i=1$ when $\bm{g}_i \neq \varnothing$ and 0 otherwise. Next, we randomly sample a subset $\bm{c}_i$ that remain consistent across varied frames from each group $\bm{g}_i$, which forms our representation for contact points $\bm{C}=\{\bm{c}_1, \bm{c}_2, \dots, \bm{c}_N | \bm{c}_i \subset \bm{g}_i\}$.

\subsection{Contact-aware Human Dynamics Branch} 
\label{sec:branch_h}

The human dynamics branch is aimed at capturing the articulated human dynamics. To achieve this, we utilize a predictor to comprehend and generate the human motions. As illustrated in Figure \ref{fig:CCL} (a), the predictor reconstructs the clean human motion and contact from noisy input based on historical conditions. Specifically, at each diffusion timestep $t$, the predictor $\mathcal{D}_H$ receives the noised vector of concatenated human motion $\bm{h}_t$ and contact $\bm{C}_t$ as input. These vectors along with timestep $t$ are embedded and then processed through an $L$-layer transformer decoder, with each block consisting of a self-attention layer, cross-attention layer and feed-forward layer. In the cross-attention layer, the features from the preceding self-attention layer serve as query, while the encoded historical conditions are used as key and value. Subsequently, we apply a linear layer to process the output feature of the predictor and directly generate the clean human motion $\bm{h}_0$ and contact $\bm{C}_0$. The loss function for human dynamics branch is formulated as:
\begin{equation}
\label{eq:L_H}
    L_{human} = ||\mathcal{D}_H([\bm{h}_t, \bm{C}_t], t, \mathcal{E}_H(\bm{y}_H)) - [\bm{h}_0, \bm{C}_0]||^2,
\end{equation}
where $[\cdot, \cdot]$ is concatenation operation. 
$\bm{y}_H$ is the provided conditions for human dynamics modeling, including historical motion sequence $\mathcal{S}^{1:T_p}$ and contact $\bm{C}^{1:T_p}$. $\mathcal{E}_H$ is a Transformer encoder that encodes conditions $\bm{y}_H$.

\subsection{Contact-aware Object Dynamics Branch}
\label{sec:branch_o}

The contact-aware object dynamics branch is developed to capture object motions with rigid translations and rotations. 
Given noised object motion $\bm{o}_t$ at diffusion timestep $t$, it employs an $L$-layer transformer predictor $\mathcal{D}_O$ with a linear projector to generate the clean object motion representation $\bm{o}_0$, as shown in Figure \ref{fig:CCL} (b). 
Specially, in each block of the predictor, we apply two cross-attention layers to incorporate the history and contact information, respectively. For the first cross-attention, we take the history embedding processed by a condition encoder as key and value to integrate the history information. 
For the second cross-attention, in order to inject contact information, we first process multiple contact points through self-attention together with a set of learnable tokens. The outputs of learnable tokens, which contain rich contextual contact information, serve as key and value of cross-attention to facilitate object motion prediction.

We use following loss function to optimize the predictor:
\begin{equation}
\label{eq:L_O}
    L_{object} = ||\mathcal{D}_O(\bm{o}_t, t,\bm{C}^{1:T_p},\mathcal{E}_O(\bm{y}_O)) - \bm{o}_0||^2,
\end{equation}
where $\mathcal{E}_O$ is the Transformer encoder that encodes the conditions $\bm{y}_O$, which includes the historical motion sequence $\mathcal{S}^{1:T_p}$ and object shape embedding.




\begin{table}[!t] 
\small
\renewcommand{\arraystretch}{1.1}
    \centering
    \caption{Quantitative comparisons on the BEHAVE dataset \cite{bhatnagar2022behave}. Text in \textbf{bold} denotes the best results. Our method significantly outperforms other approaches.}
    \vspace{-1mm}
    \begin{tabular}{c *{4}{c@{\hspace{7.3pt}}}}
        \toprule[2pt]
        \multirow{2}*{Method}   & \multicolumn{4}{c}{BEHAVE Dataset}   \\
        \cline{2-5}
         & MPJPE-H ↓    & Trans. Err. ↓    & Rot. Err. ↓    & Pene. ↓ \\
        \hline
        InterRNN       & 165     & 139   & 267     & 314        \\
        InterVAE       & 145     & 125   & 268     & 222        \\
        InterDiff      & 140     & 123   & 226     & 164        \\
        Ours        & \textbf{123}     & \textbf{106}   & \textbf{200}     & \textbf{94}        \\
        \bottomrule[2pt]
    \end{tabular}
    \label{tab:BEHAVE_ST}
\end{table}


\subsection{Contact Consistency Constraint}
\label{sec:CCC}


We introduce contact points to link the distinct human and object dynamics.
Our key insight is that contact points, which are simultaneously present on both the highly structured human body and the rigidly moving object, can be predicted by the two corresponding branches, respectively. Therefore, the consistency of these contact predictions naturally serves as a bridge between human and object dynamics.

In light of this, we formulate a consistency constraint loss to align the contact predictions across human and object branches, explicitly ensuring coherent dynamics modeling between them (refer to Figure \ref{fig:Architecture}). Formally, the contact consistency constraint is defined to penalize the difference between contact prediction in the human dynamics branch $\hat{\bm{C}}_H$ and that in the object dynamics branch $\hat{\bm{C}}_O$, expressed as:
\vspace{-1mm}
\begin{equation}
\label{eq:L_cons}
   L_{consistency} = \sum_{i=1}^{T_p+T_f} \mathcal{M}_i \odot ||\hat{\bm{C}}^i_H - \hat{\bm{C}}^i_O||^2,
   \vspace{-1mm}
\end{equation}
where $\mathcal{M}$ is a binary contact label indicating the presence (1) or absence (0) of the contact points in the corresponding regions. $\hat{\bm{C}}^i_H$ and $\hat{\bm{C}}^i_O$ refer to the contact predictions for the $i$-th frame in the human and object dynamics branches. 

The contact prediction $\hat{\bm{C}}_H$ can be obtained from the output of human dynamics branch.
As for the contact prediction $\hat{\bm{C}}_O$, we calculate them based on the physical motion laws of surface points on rigid objects: $\hat{\bm{C}}_O = \hat{\bm{R}} \cdot \bm{p} + \hat{\bm{L}}_o$. Here, $\hat{\bm{R}}$ and $\hat{\bm{L}}_o$ are predicted object motion. $\bm{p}$ is the ground-truth positions of contact points on object's surface at the rest pose.




In this way, the decoupled learning for human and object dynamics can be mutually influenced and constrained through contact consistency, which facilitates more accurate human-object prediction in a coherent manner.

\begin{figure}[!t]
\centering
\vspace{-0.5mm}
\includegraphics[width=\linewidth]{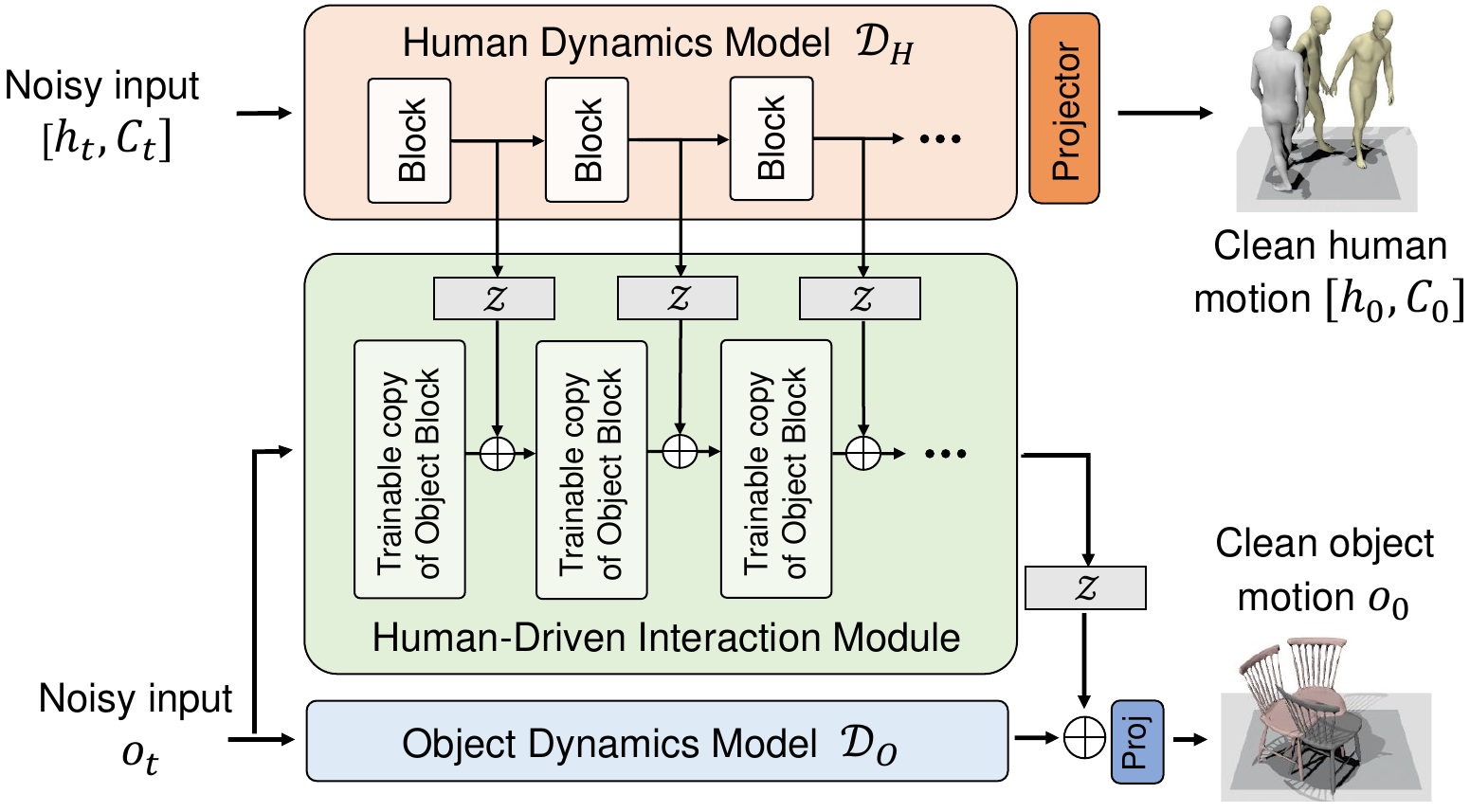}
\vspace{-5.5mm}
\caption{The Human-driven Interaction Module (HIM). The modules $\mathcal{Z}$ in gray represent the Fully-Connected (FC) layers whose weights and bias are initialized to zeros.}
\label{fig:HIM}
\end{figure}

\subsection{Human-Driven Interaction Module}
\label{subsec:CSIM}


In the context of HOI, human typically take the dominant role as force-exerting agent on objects. Inspired by this, we introduce a human-driven interaction module that propagates human dynamics to object dynamics, aiming to further enhance the coherent human-object dynamics modeling.

Our Human-driven Interaction Module (HIM) is built upon the pretrained human and object dynamics branches (training details are elaborated in Sec. \ref{sec:training}). As shown in Figure \ref{fig:HIM}, the HIM is constructed as the trainable copy of object dynamics model. We regard human dynamics as conditional control, and inject it into the intermediate feature of each HIM block, with an FC layer $\mathcal{Z}$ for connection. 
Following the practice in ControlNet \cite{zhang2023adding}, we initialize both the weight and bias in $\mathcal{Z}$ to zeros. Finally, the HIM output features are passed through an FC layer and subsequently fused with the object branch features to refine the object dynamics.

\subsection{Model Training}
\label{sec:training}

We train our model with a three-stage optimization strategy. 
In the first stage, we turn off the HIM and exclusively train the human and object dynamics branches. In the second stage, we freeze the human and object dynamics branches and only optimize the HIM module. Finally, we jointly finetune the whole system in the third stage. All stages use the following loss function for optimization:
\begin{equation}
\label{eq:L_all}
   L_{all} = \lambda_H L_{human} + \lambda_O L_{object} + \lambda_C L_{consistency},
\end{equation}
where $\lambda_H$, $\lambda_O$ and $\lambda_C$ are weights to control the trade-off between different loss terms.

\section{Experiment}
\label{sec:exp}

In this section, we conduct extensive experiments and the results show that our approach consistently outperforms the state-of-the-art methods both quantitatively and qualitatively. Furthermore, ablation studies are provided to demonstrate the effectiveness of key designs in our framework.

\begin{table}[!t] 
\small
\renewcommand{\arraystretch}{1.1}
    \centering
    \caption{Quantitative comparisons on the Human-Object Interaction dataset \cite{wan2022learn}. We evaluate the model on the test set that was not seen during training. Text in \textbf{bold} denotes the best results.}
    \vspace{-1mm}
    \begin{tabular}{*{4}{c@{\hspace{7pt}}} c @{\hspace{2pt}}}
        \toprule[2pt]
        Method   & MPJPE-H ↓    &  MPJPE-O ↓     & Trans. Err. ↓    & Rot. Err. ↓    \\
        \hline
        HO-GCN         & 111     & 153    & 123     & 303    \\
        CAHMP          & 111     & 132    & 111     & 164    \\
        InterRNN        & 124     & 127    & 109     & 151    \\
        InterVAE        & 108     & 125    & 100     & 178    \\
        InterDiff       & 105     & 84     & 60      & 120    \\
        Ours               & \textbf{100}     & \textbf{83}     & \textbf{58}      & \textbf{118}  \\
        \bottomrule[2pt]
    \end{tabular}
    \label{tab:HOI_ST}
\end{table}

\begin{table}[!t] 
\small
\renewcommand{\arraystretch}{1.1}
    \centering
    \caption{Quantitative analysis of key designs on BEHAVE. We investigate the effectiveness of Decoupled human-object dynamics modeling (Dcp), Contact injection (Cont) to dual-branch diffusion, Contact Consistency Constraint (CCC) and Human-driven Interaction Module (HIM). We report the results of InterDiff \cite{xu2023interdiff} as our Baseline.}
    \vspace{-1mm}
    \begin{tabular}{@{\hspace{2pt}}c@{\hspace{3pt}}|@{\hspace{3pt}}c@{\hspace{3pt}}c@{\hspace{3pt}}c@{\hspace{3pt}}c@{\hspace{3pt}}|@{\hspace{3pt}} c @{\hspace{3pt}} c @{\hspace{7pt}} c @{\hspace{7pt}} c @{\hspace{1pt}}}
        \toprule[2pt]
        Method  & Dcp & Cont  & CCC   & HIM    & \makecell[c]{MPJPE\\-H}↓     & \makecell[c]{Trans.\\Err.}↓    & \makecell[c]{Rot.\\Err.}↓    & Pene.↓ \\
        \hline
        Baseline   & \ding{55}   & \ding{55}   & \ding{55}   & \ding{55}       & 140     & 123   & 226     & 164        \\
        \hline
        \multirow{4}*{Ours}   & \checkmark  & \ding{55}  & \ding{55}   & \ding{55}   & 133     & 115   & 206     & 103        \\
                   & \checkmark   & \checkmark   & \ding{55}   & \ding{55}   & 130     & 112   & 205     & 97        \\
                   & \checkmark   & \checkmark  & \checkmark   & \ding{55}   & 126     & 108   & 204     & 96        \\
                   & \checkmark   & \checkmark  & \checkmark   & \checkmark   & \textbf{123}     & \textbf{106}   & \textbf{200}     & \textbf{94} \\
        \bottomrule[2pt]
    \end{tabular}
    \label{tab:Ablation}
\end{table}

\subsection{Experimental Setup}

\noindent\textbf{Datasets.} 
Our method is evaluated on two 3D human-object interaction benchmark datasets: BEHAVE \cite{bhatnagar2022behave} and Human-Object Interaction \cite{wan2022learn} datasets. The BEHAVE is a large-scale dataset that records around 15.2k frames. It contains interaction sequences of 8 subjects with 20 different 3D objects, with a framerate at 30 Hz. As in \cite{xu2023interdiff}, human poses are represented using the SMPL-H \cite{shafir2023human,mandery2015kit} parameter, and we denote the object poses using 6D rotation and translation. The Human-Object Interaction dataset involves recordings of 6 subjects interacting with 12 objects from 6 categories. Following \cite{xu2023interdiff}, we represent the human pose by a 21-joint skeleton, and 12 key points for the objects. For data splitting, 17,770 sequences are used for training (seen) and 582 sequences for testing (unseen).

\begin{table}[!t] 
\small
\renewcommand{\arraystretch}{1.1}
    \centering
    \caption{Analysis of the asymmetric structure on BEHAVE dataset. CoopDiff with asymmetric structure outperforms human-aligned and object-aligned symmetric variants.}
    \vspace{-1mm}
    \begin{tabular}{@{\hspace{3pt}}c@{\hspace{4pt}}|@{\hspace{3pt}}c @{\hspace{6pt}} c @{\hspace{6pt}} c @{\hspace{6pt}} c @{\hspace{1pt}}}
        \toprule[2pt]
        Method   & MPJPE-H ↓    & Trans. Err. ↓    & Rot. Err. ↓    & Pene. ↓ \\
        \hline
        Human-aligned     & 126     & 108   & 203     & 96        \\
        Object-aligned       & 125     & 107   & 203     & 95        \\
        Ours (CoopDiff)    & \textbf{123}     & \textbf{106}   & \textbf{200}     & \textbf{94} \\
        \bottomrule[2pt]
    \end{tabular}
    \label{tab:abla_asym}
\end{table}

\noindent\textbf{Evaluation Metrics.}
For performance evaluations, we employ the same metrics and configurations as \cite{xu2023interdiff}. Specifically, MPJPE-H metric measures the average $l_2$ distance between ground truth (GT) and predicted human joint positions. To assess object motion accuracy, we report MPJPE-O, Trans. Err and Rot. Err as the average $l_2$ distance between GT and predicted object key points, translations and quaternions, respectively. The Pene. indicates the average percentage of object vertices with non-negative values in the human signed distance function \cite{petrov2023object}.


\begin{figure}[t]
\centering
\includegraphics[width=\linewidth]{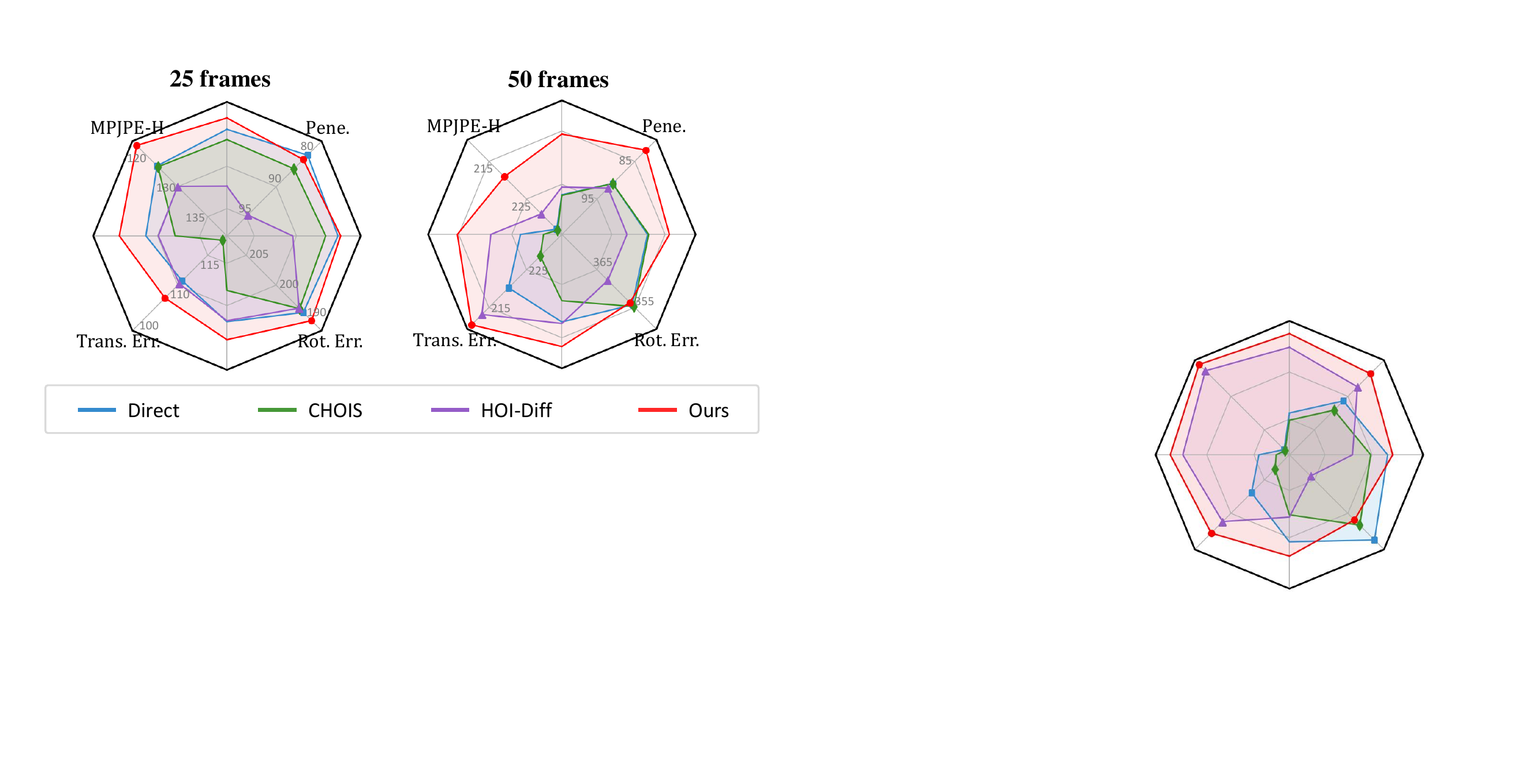}
\vspace{-5mm}
\caption{Comparisons of our contact consistency modeling against other contact-guided approaches, including CHOIS \cite{li2023controllable}, HOI-Diff \cite{peng2025hoi}, and direct alignment of the nearest human-object points\protect\footnotemark.}
\label{fig:Abla_cont}
\end{figure}

\footnotetext{In the direct-alignment approach, contact is determined as the nearest point on the human mesh to the object, and the object surface is directly moved to this point during post-processing.}

\begin{figure*}[!t]
\centering
\vspace{-0.6mm}
\includegraphics[width=0.87\linewidth]{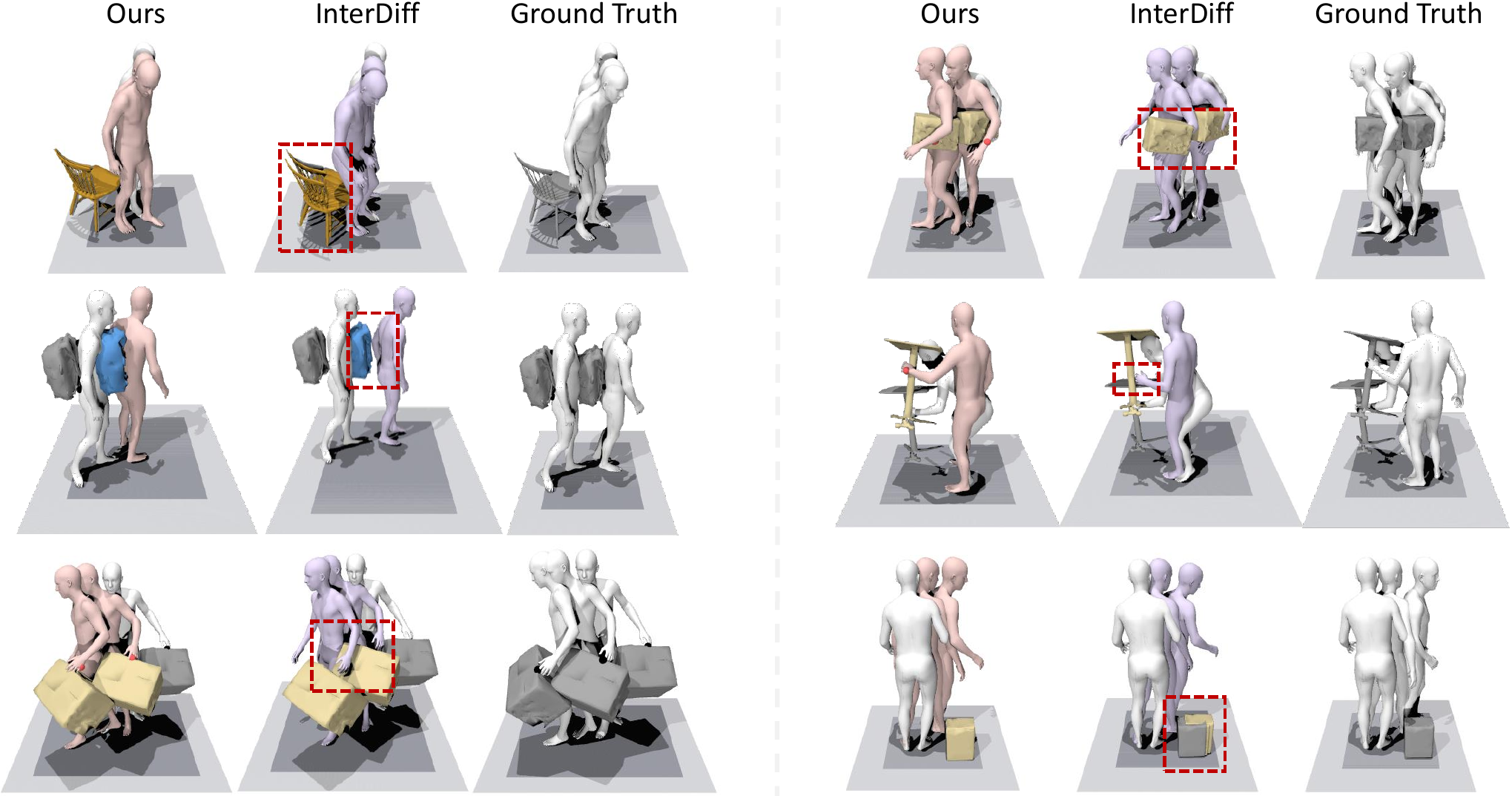}
\vspace{-1.5mm}
\caption{Visualization of future interactions predicted by our method against InterDiff on the BEHAVE dataset. The results produced by InterDiff contain some unrealistic interactions (highlighted in red boxes), such as interpenetration, object floating artifacts, and moving objects in the absence of contact. In contrast, our CoopDiff generates more accurate and realistic interactions. The history and ground-truth HOIs are in gray and the predictions are in color. Best viewed in color.}
\label{fig:BEHAVE_vis_ST}
\end{figure*}

\begin{figure}[!t]
\centering
\vspace{-0.3mm}
\includegraphics[width=0.85\linewidth]{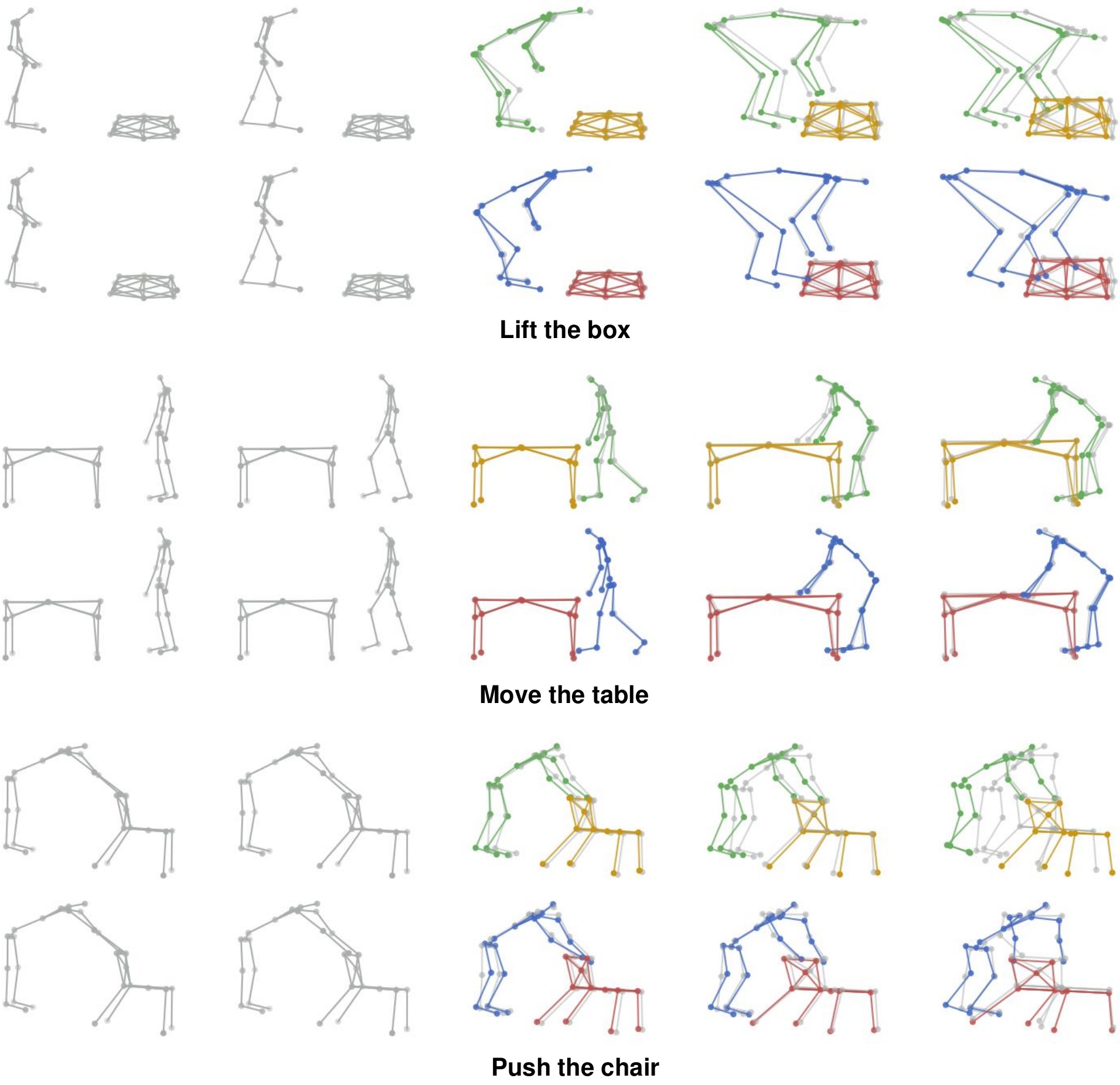}
\vspace{-2mm}
\caption{Visualization comparisons on the Human-Object Interaction dataset. The human and object motions predicted by our CoopDiff and InterDiff are in blue and red, green and yellow, respectively. The ground truth (GT) is in gray.}
\vspace{-1mm}
\label{fig:HOI_pred}
\end{figure}

\noindent\textbf{Benchmark methods.}
Despite the significance of 3D HOI anticipation, the task of predicting complex interactions with dynamic objects still remains underexplored. Therefore, only limited prior research is currently available for direct use as baselines. We list them as follows: (i) \textbf{InterDiff} \cite{xu2023interdiff}, which jointly predicts human and object dynamics by a diffusion model, with a correction interaction module to refine predictions; (ii) \textbf{InterVAE} \cite{xu2023interdiff,petrovich2022temos} extends Transformer-based VAE models to this setting; (iii) \textbf{InterRNN} \cite{xu2023interdiff,rempe2020predicting} employs an LSTM-based model for prediction. To test generalization on skeleton-based HOI datasets, (iv) \textbf{CAHMP} \cite{corona2020context} and (v) \textbf{HO-GCN} \cite{wan2022learn} are also included for comparisons.

\noindent\textbf{Implementation Details.}
In our implementation, both human and object branches use 8 Transformer layers \cite{vaswani2017attention} for condition encoder and 8 for dynamics decoder, where the dimension is set to 256, and 4 attention heads are applied. We follow \cite{xu2023interdiff} by adopting interaction correction module to refine human and object predictions. As for Human-Object Interaction dataset where only skeletal representations of humans and objects are available, we define contact points as the human joints that are closest to the object. All experiments are conducted on RTX 3090 GPUs.


\begin{figure*}[!t]
\centering
\vspace{-1mm}
\includegraphics[width=0.9\linewidth]{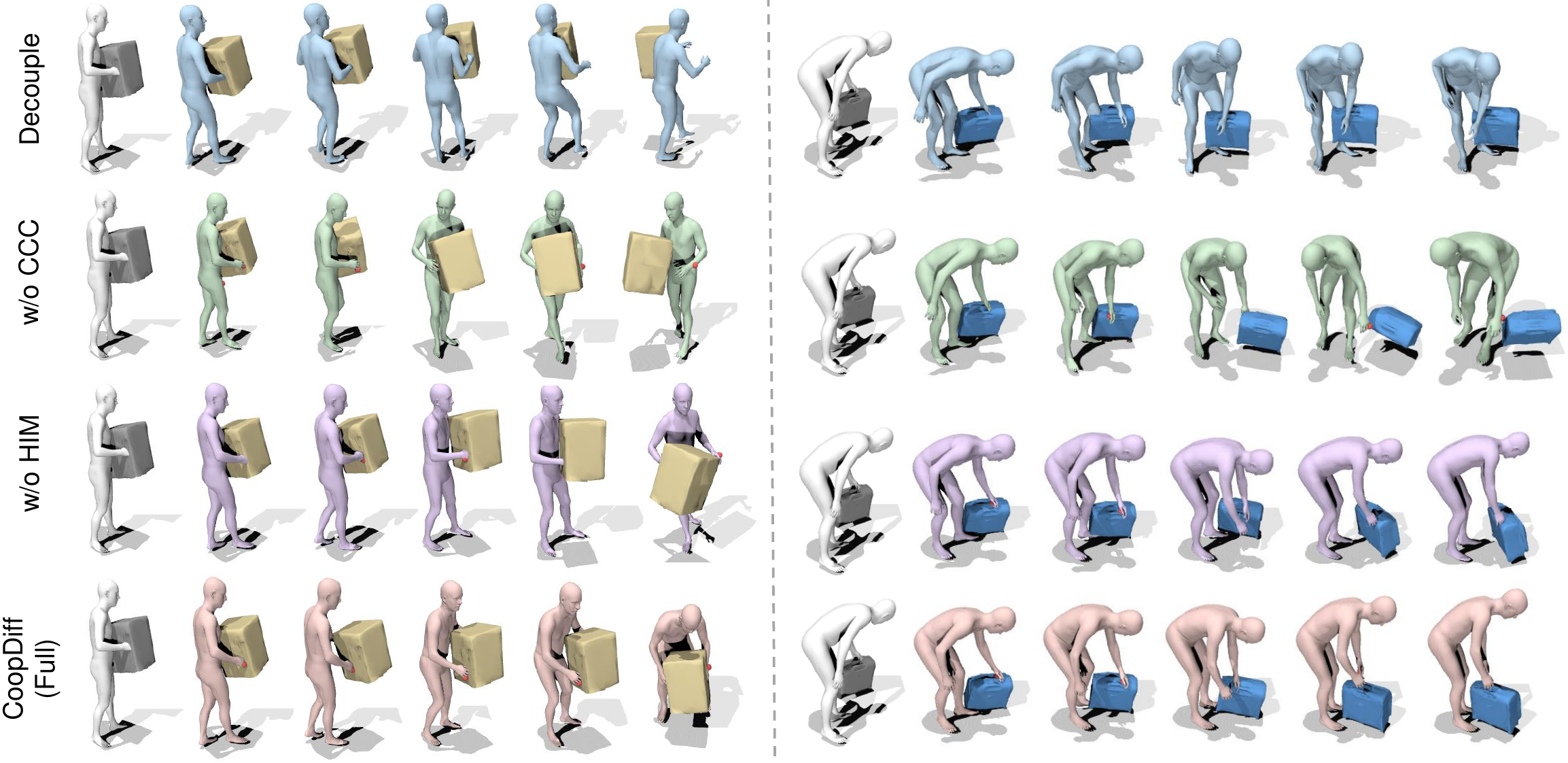}
\vspace{-2.2mm}
\caption{Visual analysis on mitigating unrealistic interactions. With our contact-consistent modeling, humans and objects exhibit more coherent dynamics near the contact regions. The HIM module further promotes realistic interactions, which reduces slight penetration and facilitates achieving authentic contact. The red dots denote visible contact points outputted by CoopDiff.}
\label{fig:ablation}
\end{figure*}

\subsection{Comparisons with State-of-the-Art Methods}
\label{subsec:comp}

We quantitatively compare CoopDiff with other methods on BEHAVE \cite{bhatnagar2022behave} and Human-Object Interaction \cite{wan2022learn} datasets. The results suggest that our method consistently achieves state-of-the-art performance under various experimental settings across both datasets.

\noindent\textbf{BEHAVE Dataset.} 
We first compare our CoopDiff with existing state-of-the-art on the BEHAVE dataset. As shown in Table \ref{tab:BEHAVE_ST}, our method significantly outperforms InterDiff \cite{xu2023interdiff}, improving all the metrics by over 12.1\%, 13.8\%, 11.5\% and 43\% in MPJPE-H, Trans. Err., Rot. Err. and Pene. metrics, respectively. These results demonstrate the effectiveness of our framework in accurately capturing the distinct motion patterns of humans and objects, while reducing interpenetration in HOI modeling (Pene. metric).





\noindent\textbf{Human-Object Interaction Dataset.} We additionally test our approach on the Human-Object Interaction dataset, focusing on a challenging scenario that evaluates instances unseen during training.
Table \ref{tab:HOI_ST} presents the prediction performance of our CoopDiff against various methods on the Human-Object Interaction dataset. As shown, our CoopDiff again achieves the best results across all evaluation metrics. 

\subsection{Ablation Study}


We further conduct ablation studies on BEHAVE dataset to comprehensively analyze different components in CoopDiff.


\noindent\textbf{Analysis on the key designs.}
To assess the impact of key designs on the system performance, we begin with the Baseline model (InterDiff \cite{xu2023interdiff}), then sequentially decouple human-object dynamics modeling, integrate contact points into both branches, enforce contact consistency constraint, and incorporate the human-driven interaction module. As in Table \ref{tab:Ablation}, each component contributes positively to improving the system performance across all metrics. 

\noindent\textbf{Effectiveness of our contact modeling.}
Contact points are essential for achieving realistic interactions in mesh-based HOI generation. Beyond prediction tasks, other methods also incorporate contact information: HOI-Diff~\cite{peng2025hoi} devises a specialized contact predictor and aligns human-object motions based on the predicted contact; CHOIS \cite{li2023controllable} leverages classifier guidance to ensure the contact between hand and object. However, these methods use contact primarily for post-processing, leading to an underutilization of contact information. In Figure \ref{fig:Abla_cont}, we replace our contact modeling with theirs and compare the prediction performance. As shown, our contact guidance achieves the highest accuracy and minimal penetration, indicating the superiority of our contact modeling over alternative approaches for accurate and coherent HOI prediction.



\noindent\textbf{Effectiveness of the structure of human-object branches.}
In the human branch, we integrate historical contact information with the condition encoder; while in object branch, we additionally use a set of learnable tokens for early aggregation.
We validate this asymmetric design against two variants: i) both branches using learnable tokens, and ii) both using only the condition encoder. Table \ref{tab:abla_asym} empirically shows that the asymmetric design yields better performance than the others. 
We speculate the reason is that: the object representation is relatively simple, involving only the centroid and rotation, hence early aggregation of complicated historical contacts helps reduce noise and stabilizes learning.
In contrast, the human representation involves motion of multiple joints, and incorporating the raw contact sequence may help capture complicated historical interactions. 
Therefore, we keep the asymmetric structure in CoopDiff by default.

\subsection{Qualitative Analysis}

\noindent\textbf{Qualitative comparisons.}
Figure \ref{fig:BEHAVE_vis_ST} visualizes future interactions predicted by our method against InterDiff \cite{xu2023interdiff} on the BEHAVE dataset. As shown, our CoopDiff generates interactions that align more closely with GT and appear more realistic, with fewer cases of interpenetration, object floating and moving object without contact compared to InterDiff. We also present predictions on the Human-Object Interaction dataset in Figure \ref{fig:HOI_pred}. The results further indicate the superiority of our approach over InterDiff, even under the unseen scenarios in training data. More visualization results (videos) are provided in the supplementary.


\noindent\textbf{Qualitative results on realistic interactions.} 
Our framework with contact integration, contact-consistent constraint and HIM module can effectively mitigate unrealistic artifacts. Figure \ref{fig:ablation} presents some qualitative results. We observe that: i) incorporating contact points enables the model to perceive human-object interaction locations, alleviating inconsistency between human and object motion; ii) with contact consistent constraint, humans and objects exhibit more coherent dynamics near contact regions, e.g., objects transition from floating states to a position close to hand; iii) HIM promotes realistic interactions, which reduces penetration and facilitates achieving authentic contact. These results demonstrate the effectiveness of our method and modular designs.


\section{Conclusion}
\label{sec:conclusion}

We present CoopDiff, a novel contact-consistent decoupled diffusion framework for HOI anticipation. CoopDiff mainly comprises a human dynamics branch that models highly structured human dynamics, and an object dynamics branch that captures object motion with rigid translations and rotations. 
To accommodate the distinct dynamics between human and object, we introduce contact as shared anchors and feed them into both branches for additional contact predictions. These predictions are then enforced with a consistency constraint for coherent dynamics modeling. To further enhance alignment, we devise human-driven interaction module, considering that humans always take the dominant role as force-exerting agents on objects. Extensive experiments demonstrate that our CoopDiff achieves SOTA performance.

%



\bibliography{aaai2026}



\end{document}